\documentclass[a4paper, oneside, twocolumn, notitlepage, 10pt]{extarticle_ecoc}
\usepackage{ecoc}
\usepackage{soul}
\usepackage{tabularx}
\usepackage{tikz}
\usetikzlibrary{shapes.geometric, arrows.meta, positioning, fit, calc}
\usepackage{pgfplots}
\pgfplotsset{compat=1.18}

\begin{document}
\selectlanguage{english}    


\title{Hybrid Active-Online Learning Framework for Label-Efficient Concept Drift Adaptation in Optical Network Failure Detection}%


\author{
    Yousuf Moiz Ali\textsuperscript{(1)},
    Jaroslaw E. Prilepsky\textsuperscript{(1)},
    Jo{\~a}o Pedro\textsuperscript{(2)},
Sasipim Srivallapanondh\textsuperscript{(3)},\\
    Antonio Napoli\textsuperscript{(3)},
    Sergei K. Turitsyn\textsuperscript{(1)},
    Pedro Freire\textsuperscript{(1)}
}

\maketitle                  


\begin{strip}
    \begin{author_descr}

\textsuperscript{(1)}~Aston University, Birmingham, UK,
   \textcolor{blue}{\uline{y.moizali@aston.ac.uk}},
   \textsuperscript{(2)}~Nokia, Optical Networks, Carnaxide, Portugal, 
   \textsuperscript{(3)}~Nokia, Munich, Germany

    \end{author_descr}
    \vspace{-2mm}
\end{strip}

\renewcommand\footnotemark{}
\renewcommand\footnoterule{}


\begin{strip}
    \begin{ecoc_abstract}
        We propose a hybrid active-online learning framework for label-efficient concept drift adaptation in optical network failure detection. Using margin-based selective labeling, our method achieves near-ceiling accuracy and AUC scores while querying only 3.4\% of streaming samples, with negligible latency overhead compared to static inference. ©2026 The Author(s) 
    \end{ecoc_abstract}
    \vspace{-3.0mm}
\end{strip}

\vspace{-2.0mm}
\section{Introduction}
\vspace{-1.0mm}
Effective and timely failure management is of utmost importance in optical networks to ensure continuous service to end users~\cite{moiz2025data}. For this task, different machine learning (ML) based methods have been proposed to automate the entire failure management pipeline and reduce the risk of human error~\cite{moiz2025data, musumeci2025failure, musumeci2019tutorial}. However, these ML models often struggle to adapt when the data distribution changes~\cite{shayesteh2022automated}. In an optical network, the distribution of the data points can change due to various factors such as equipment aging or varying channel properties \cite{andre2005chromatic,bebawi2018comprehensive}. This shift in the data's statistical properties is referred to as Concept Drift (CD)~\cite{lu2018learning}. Static models, which remain fixed after training, generally fail to adapt to CD in the dataset, thereby leading to misclassifications.

\begin{figure*}[b!]
\vspace{-6mm}
  \centering
  \includegraphics[width=0.75\textwidth]{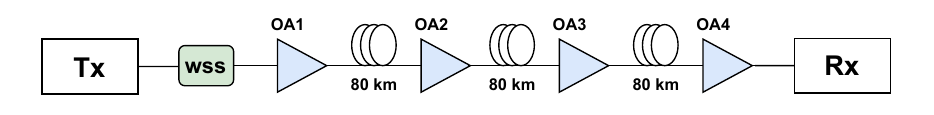}
  \captionsetup{font=footnotesize}
\caption{Experimental testbed setup used to generate the dataset. The Wavelength Selective Switch (WSS) was used to introduce attenuation at OA1 to simulate normal and failure conditions.}
\vspace{-2mm}
\label{fig:testbed}
\end{figure*}

One prominent example of CD in optical telemetry occurs when a model trained on soft failure data encounters hard failure events. Hard failures are rare and difficult to simulate, given the limited data from network operators and vendors. Online learning has emerged as one of the tools for adapting to CD. Online learning differs from traditional batch learning in that data arrives sequentially, and the model is updated once per sample. Training online models is prequential: the model first makes a prediction on a new sample, then learns from it~\cite{hoi2021online}.

In our recent work~\cite{ali2026experimental}, we proposed an online learning method for failure detection in optical networks. This approach was adapted to new hard failures and maintained low latency. However, because the models were trained using fully supervised learning, we assumed labels were constantly available. In real-world installations, this assumption rarely holds. Labels are often missing during training because experts or automated oracles are not always available to provide them. Active learning addresses this challenge by selectively querying only the most informative samples, making it suitable for label-scarce settings~\cite{settles2009active}.


Active learning has received some attention in optical networks literature, but all works focused on Quality of Transmission (QoT) estimation~\cite{li2021predictive, azzimonti2020comparison, azzimonti2019reducing, azzimonti2020active, azzimonti2019using}. To this end, we propose a novel hybrid active-online learning approach for failure detection in optical networks that leverages the power of online learning to adapt to CD in the dataset while using only the most informative samples, rather than the entire dataset, via active learning. The novelty of this work lies, to the best of our knowledge, in the first ever joint usage of active and online learning approaches, thus providing the best of both worlds. Our results indicate that the hybrid approach achieves the same accuracy and Area-Under-the-Curve (AUC) score as the supervised model while only using 3.4\% of the available samples. The latency overhead of the hybrid approach is also very minimal, with only a 0.0026~ms increase in per-event latency.

\vspace{-2mm}
\section{Methodology and Dataset}
In this work, we used the dataset from~\cite{silva2022learning}. Fig.~\ref{fig:testbed} shows the testbed used to generate the data. The dataset is split into separate soft-failure (SFD) and hard-failure (HFD) datasets. It includes four features: Bit-Error-Rate (BER) and Optical Signal-to-Noise Ratio (OSNR) measured at both the transmitter (\textit{BER\_Tx}, \textit{OSNR\_Tx}) and receiver (\textit{BER\_Rx}, \textit{OSNR\_Rx}).

\begin{figure*}[t!]
\vspace{-6mm}
  \centering
  \includegraphics[width=\textwidth]{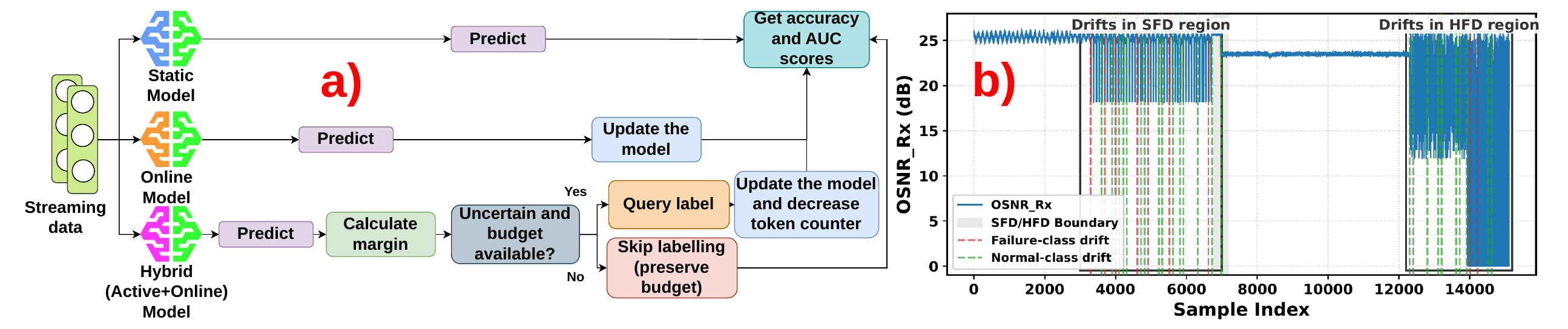}
  \captionsetup{font=footnotesize}
\caption{a) System design for the static, supervised online, and hybrid (active + online) systems. All three models were pre-trained on the SFD, and the HFD was used as streaming data. b) Data distribution of the \textit{OSNR\_Rx} feature, separated by the SFD (left) and HFD (right) boundary. Confirmed drifts are marked green (normal) and red (failure).}
\vspace{-2mm}
\label{fig:system+osnr}
\end{figure*}

Fig.~\ref{fig:system+osnr}a shows the system design for the static, supervised online, and the hybrid system. The static model is used to predict only on the new sample. The supervised online model is trained in a prequential manner: it first predicts on the new sample, then learns from it. The hybrid model first predicts on the new sample, and then we use Uncertainty Sampling~\cite{settles2009active} to calculate the difference between the probabilities of the sample being a failure or normal. If the difference exceeds the \textit{Margin Threshold} (tunable parameter) and there is budget to query the sample, we query it and obtain the label; otherwise, we skip the sample entirely. If the sample was queried, the hybrid model is updated with that sample, and one token is deducted from the budget counter. The budget increases by one token after a set number of steps, until it reaches the maximum allowed value. Uncertainty Sampling ensures budget is reserved only for the most informative samples and filter out those that might not add value to its parameters.

To simulate real-world telemetry, we pre-trained all three models on the SFD and used the HFD as streaming data. To identify CD in the dataset, a two-stage procedure was deployed: the Page-Hinkley Test~\cite{page1954continuous, hinkley1970} (PHT) as the primary check, followed by the Kolmogorov-Smirnov (KS) test~\cite{kstest} as the second validation layer. PHT computes the running minimum and compares it with the cumulative sum of the data points; a drift is signaled if the value exceeds a set threshold. The KS test compares the cumulative distributions of two windows of data points. Currently, the drift detection mechanism is solely used for detecting drifts in the dataset and for visualization purposes and has no effect on the budget and the \textit{Margin Threshold}.

\begin{figure}[b!]
\vspace{-5mm}
  \centering
  \includegraphics[width=\columnwidth]{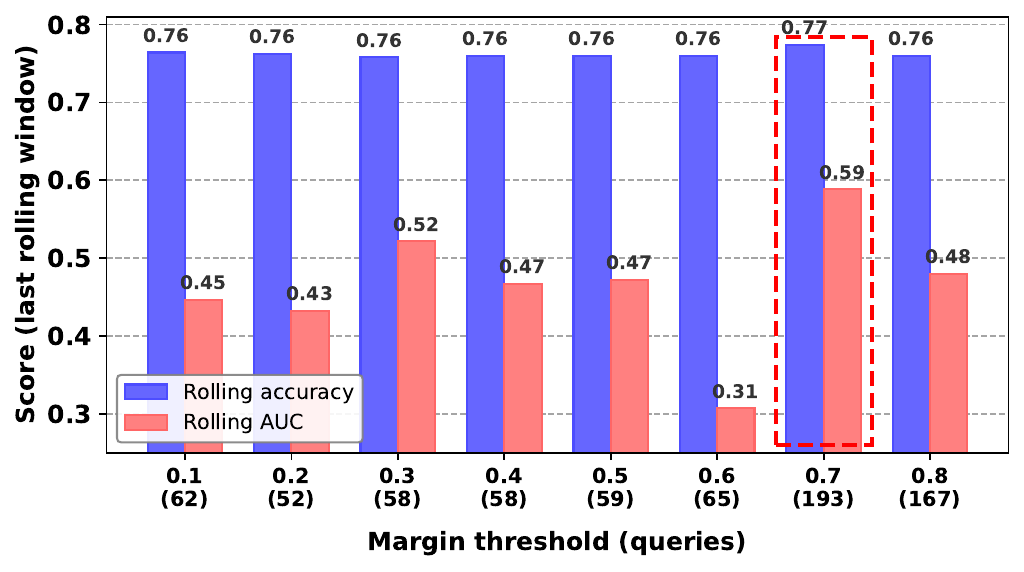}
  \captionsetup{font=footnotesize}
  \caption{Margin threshold impact on hybrid performance. Values show the final rolling window (500 samples) with query counts in parentheses. The red dashed box indicates the optimal threshold.}
  \label{fig:threshold_sweep}
\end{figure}


Fig.~\ref{fig:system+osnr}b shows data distribution for the \textit{OSNR\_Rx} feature, and the confirmed drifts in the SFD and the HFD. We are only considering drifts in the \textit{OSNR\_Rx} feature; it exhibited the highest correlation with the failure label as it captures cumulative signal degradation across the transmission span. The difference between the drifts in the two datasets is that the failure class drifts in the HFD occur suddenly without any warning, while they build up in the SFD region. To observe the effect of extra failures, we added synthetic failure samples to the end of the HFD stream. These were generated by adding small amounts of noise to randomly selected existing failure samples.

We used the Adaptive Random Forest (ARF)~\cite{gomes2017adaptive} algorithm for the static, supervised online, and hybrid systems. We selected this ensemble model because it performs well across most datasets and is specifically designed for online learning tasks. In the static model, the ARF behaves like a standard Random Forest algorithm. The metrics used in this work are the rolling accuracy and the AUC scores, both calculated over windows of 500 data points. The ARF and hybrid system parameters, determined via trial and error, are shown in Tab.~\ref{tab:table1}.
\vspace{-3.0mm}
\begin{table}[h!]
\footnotesize
    \centering
    \caption{ML model and hybrid system parameters} \label{tab:table1}
    \begin{tabularx}{\columnwidth}{|X|X|} 
        \hline  \textbf{Model/Parameter} & \textbf{Value}                 \\
        \hline  ARF & Number of trees: 10; Max features considered when splitting: 2; Samples before split: 50                \\
        \hline  Budget Tokens (initial starting budget) & 500 \\
        \hline  Maximum Tokens & 1000 \\
        \hline  Refill Interval & 50 \\
        \hline  Margin Threshold & 0.7 \\
        \hline
    \end{tabularx}
    \vspace{-2.0mm}
\end{table}

\vspace{-4.0mm}
\section{Results and Discussion}
The first step was to identify the \textit{Margin Threshold} that balances performance with labeling cost. Fig.~\ref{fig:threshold_sweep} shows the rolling accuracy and AUC from the final window of the stream for each threshold, with query counts shown in parentheses. A threshold of 0.7 yields the highest accuracy (0.77) and AUC (0.59) while using only 193 queries (2.4\% of the original HFD stream), making it the optimal operating point for subsequent experiments.

\begin{figure*}[t!]
  \centering
  \includegraphics[width=\textwidth]{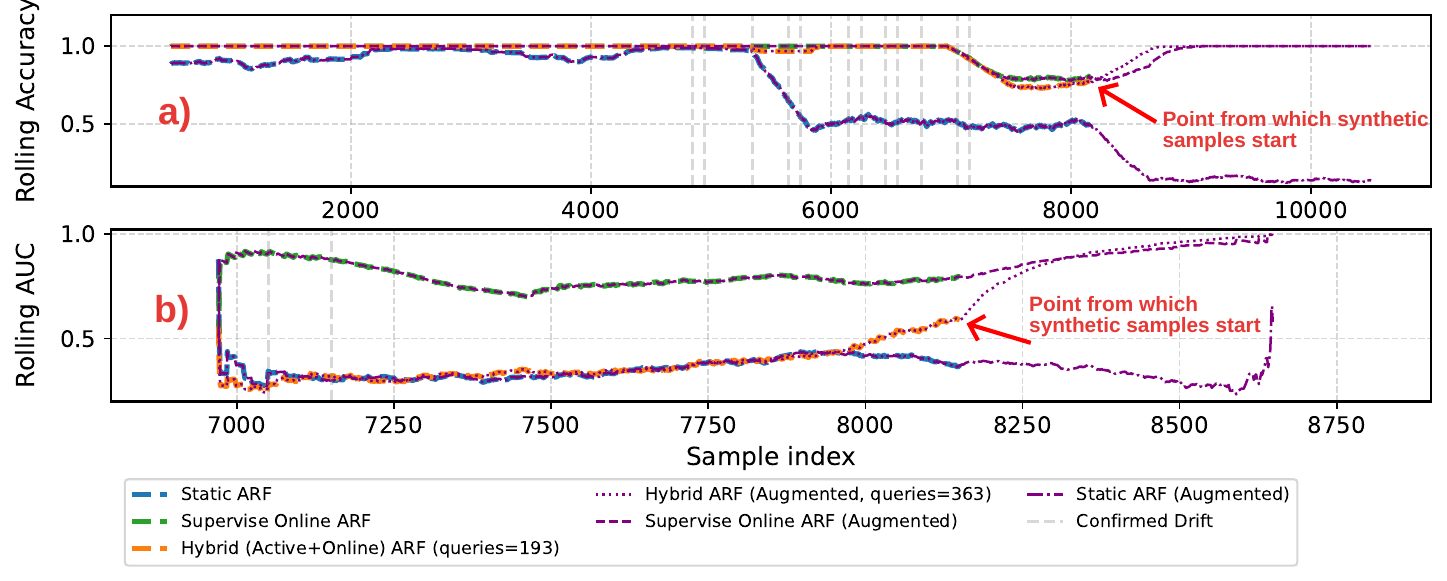}
  \captionsetup{font=footnotesize}
  \caption{a) and b) Rolling accuracy and AUC score plot on the HFD. The gray dotted lines represent the confirmed drift lines in the \textit{OSNR\_Rx} feature. The arrow indicates the point at which the synthetic samples begin.}
  \label{fig:rolling_metric}
\vspace{-2.0mm}

\end{figure*}

Fig.~\ref{fig:rolling_metric}a and Fig.~\ref{fig:rolling_metric}b show the rolling accuracy and AUC plots for the static, supervised online, and the hybrid models, respectively. The confirmed drifts in the HFD for the \textit{OSNR\_Rx} feature are shown in gray color. If we focus on Fig.~\ref{fig:rolling_metric}a, we can see that up to the end of the stream and before the synthetic samples start, the hybrid system's accuracy is mostly similar to the supervised online model, with only a slight dip when it encounters a drift at around 5200 sample index. When the first hard failure occurred around index 7200, the accuracy of both online models dipped before increasing slightly, with the supervised model having slightly higher accuracy than the hybrid model.  Both the online models (supervised and hybrid) achieve much higher accuracy than the static models, demonstrating the advantages of online learning during concept drift.

Looking at the accuracy results, it is clear that the hybrid system is working, and if we had more failure samples, it might even match the performance of the supervised online model. In the accuracy plot, the purple dotted lines represent the accuracy due to the synthetic failure samples. As more failure samples become available, the hybrid system approaches the performance of the supervised model and even surpasses it at one point, achieving 100\% accuracy. The hybrid model queried only 193 labels from the original HFD and 363 labels from the augmented data stream. This is well below the budget allocated for the number of labels that can be queried, indicating that the system is budget-efficient. The static model's accuracy further declined as it encountered more unseen failure samples. This result shows that the hybrid model matched supervised online performance using just 3.4\% of labels

Focusing on the AUC plot in Fig.~\ref{fig:rolling_metric}b, we can see that the supervised online ARF maintains a much higher AUC score than the hybrid and static models. The hybrid model shows an increasing trend towards the end of the stream, whereas the static model shows no improvement and even a further drop in performance. The purple lines show the AUC scores for the synthetic failure samples. We can see that as the hybrid system learns from more failure samples, the AUC score approaches that of the supervised online model, even with only 3.4\% of the available samples. The rolling AUC is computed only when both classes are present in the window. As the augmented stream approaches its end, the rolling window becomes dominated by failure samples, causing the static model's AUC to rise artificially before the metric becomes undefined and is no longer reported.



The next part of our analysis examined the latency incurred by the hybrid system. To calculate the latency, we repeated the experiment 100 times to ensure stable results and computed the mean per-event latency and the online overhead with a 95\% confidence interval.
For the static model, an event is making a prediction on a sample. The supervised online model also includes model updates. In the hybrid model, an event consists of making a prediction and using uncertainty sampling to decide whether to query and update.


\vspace{-3.0mm}
\begin{table}[h]
\centering
\caption{Mean per-event latency and the online overhead for all the ML models.}
\label{tab:latency}
\resizebox{\columnwidth}{!}{%
\begin{tabular}{|l|c|c|}
\hline
\textbf{Model} & \textbf{Mean Latency (ms)} & \textbf{Online Overhead (ms)} \\ \hline
Static                   & 0.1624 $\pm$ 0.0027 & --      \\ \hline
Supervised Online        & 0.5146 $\pm$ 0.0092 & 0.3522 $\pm$ 0.0082  \\ \hline
Hybrid & 0.1650 $\pm$ 0.0031 & 0.0027 $\pm$ 0.0034 \\ \hline
\end{tabular}%
}
\vspace{-3.0mm}
\end{table}

Analysis of the latency metrics in Tab.~\ref{tab:latency} reveals that the hybrid model’s per-event latency is nearly identical to the static model, with a marginal increase of 0.0026ms. This efficiency is driven by the reduced query load inherent in the active learning component. Furthermore, the hybrid model drastically reduces online overhead relative to the supervised online model, demonstrating its ability to sustain high adaptation performance at near-static latency levels.

\vspace{-3.0mm}
\section{Conclusions}
\vspace{-1.0mm}
We proposed a novel hybrid active-online learning approach for label-efficient concept drift adaptation in failure detection for optical networks. The results showed that the hybrid system achieved comparable performance to the fully supervised online model while querying only 3.4\% of the streaming samples for labels. The latency analysis confirmed that the hybrid system's per-event latency is nearly identical to that of the static model, with minimal online overhead compared to the fully supervised approach. These findings show the proposed method maintains supervised-level accuracy with static-level latency, providing a practical solution for real-time monitoring under labeling constraints.

\clearpage
\section{Acknowledgements}
This work has received funding from the European Commission MSCA-DN NESTOR project (G.A. 101119983). SKT acknowledges EPSRC project TRANSNET (EP/R035342/1). João Pedro acknowledges the HORIZON \text{ALLEGRO} project (G.A. 700001837) and FCT – Fundação para a Ciência e a Tecnologia, I.P., under project/support UID/50008/2025 – Instituto de Telecomunicações, with DOI identifier https://doi.org/10.54499/UID/50008/2025. Antonio Napoli and Sasipim Srivallapanondh acknowledge the EU HORIZON
SENSEI project (GA No. 101189545). For the purposes of open access, the authors have
applied a Creative Commons Attribution (CC BY)
licence to any Author Accepted Manuscript (AAM)
version arising from this submission.

\printbibliography[]

@article{musumeci2019tutorial,
  title={A tutorial on machine learning for failure management in optical networks},
  author={Musumeci, Francesco and Rottondi, Cristina and Corani, Giorgio and Shahkarami, Shahin and Cugini, Filippo and Tornatore, Massimo},
  journal={Journal of Lightwave Technology},
  volume={37},
  number={16},
  pages={4125--4139},
  year={2019},
  publisher={OSA},
  doi = {10.1109/JLT.2019.2922586}
}

@article{lu2018learning,
  title={Learning under concept drift: A review},
  author={Lu, Jie and Liu, Anjin and Dong, Fan and Gu, Feng and Gama, Joao and Zhang, Guangquan},
  journal={IEEE Transactions on Knowledge and Data Engineering},
  volume={31},
  number={12},
  pages={2346--2363},
  year={2018},
  publisher={IEEE},
  doi = {10.1109/TKDE.2018.2876857}
}

@article{hoi2021online,
  title={Online learning: A comprehensive survey},
  author={Hoi, Steven CH and Sahoo, Doyen and Lu, Jing and Zhao, Peilin},
  journal={Neurocomputing},
  volume={459},
  pages={249--289},
  year={2021},
  publisher={Elsevier},
    doi = {https://doi.org/10.1016/j.neucom.2021.04.112}
}

@inproceedings{ali2026experimental,
  title={Experimental Demonstration of Online Learning-Based Concept Drift Adaptation for Failure Detection in Optical Networks},
  author={Ali, Yousuf Moiz and Prilepsky, Jaroslaw E and Pedro, Jo{\~a}o and Napoli, Antonio and Srivallapanondh, Sasipim and Turitsyn, Sergei K and Freire, Pedro},
  booktitle={2026 Optical Fiber Communications Conference and Exhibition (OFC)},
  pages={1--3},
  organization={IEEE},
  note={available as arXiv preprint arXiv:2602.10401},
  year={2026},
  doi={https://doi.org/10.48550/arXiv.2602.10401}
}

@inproceedings{li2021predictive,
  title={Predictive uncertainty aware active learning for regression-based QoT estimation in optical networks},
  author={Li, Zheng and Gu, Zhiqun and Zhang, Jiawei and Zhou, Yuhang and Ji, Yuefeng},
  booktitle={Asia Communications and Photonics Conference},
  pages={T2B--4},
  year={2021},
  organization={Optica Publishing Group},
  doi={https://doi.org/10.1364/ACPC.2021.T2B.4}
}

@techreport{settles2009active, 
Author = {Burr Settles}, 
Institution = {University of Wisconsin--Madison},
Number = {1648}, 
Title = {Active Learning Literature Survey}, 
Type = {Computer Sciences Technical Report}, 
Year = {2009} 
}

@article{azzimonti2020comparison,
  title={Comparison of domain adaptation and active learning techniques for quality of transmission estimation with small-sized training datasets},
  author={Azzimonti, Dario and Rottondi, Cristina and Giusti, Alessandro and Tornatore, Massimo and Bianco, Andrea},
  journal={Journal of Optical Communications and Networking},
  volume={13},
  number={1},
  pages={A56--A66},
  year={2020},
  publisher={Optical Society of America},
doi = {10.1364/JOCN.401918}}

@inproceedings{azzimonti2020active,
  title={Active vs transfer learning approaches for QoT estimation with small training datasets},
  author={Azzimonti, Dario and Rottondi, Cristina and Giusti, Alessandro and Tornatore, Massimo and Bianco, Andrea},
  booktitle={Optical Fiber Communication Conference},
  pages={M4E--1},
  year={2020},
  organization={Optica Publishing Group},
  doi={https://doi.org/10.1364/OFC.2020.M4E.1}
}

@inproceedings{azzimonti2019using,
  title={Using active learning to decrease probes for QoT estimation in optical networks},
  author={Azzimonti, Dario and Rottondi, Cristina and Tornatore, Massimo},
  booktitle={Optical Fiber Communication Conference},
  pages={Th1H--1},
  year={2019},
  organization={Optica Publishing Group},
  doi={https://doi.org/10.1364/OFC.2019.Th1H.1}
}

@article{silva2022learning,
  title={Learning long-and short-term temporal patterns for ML-driven fault management in optical communication networks},
  author={Silva, Mois{\'e}s Felipe and Pacini, Alessandro and Sgambelluri, Andrea and Valcarenghi, Luca},
  journal={IEEE Transactions on Network and Service Management},
  volume={19},
  number={3},
  pages={2195--2206},
  year={2022},
  publisher={IEEE},
  doi={10.1109/TNSM.2022.3146869}
}

@article{page1954continuous,
 author = {E. S. Page},
 journal = {Biometrika},
 number = {1/2},
 pages = {100--115},
 title = {Continuous Inspection Schemes},
 volume = {41},
 year = {1954},
 doi={https://doi.org/10.2307/2333009}
}

@article{hinkley1970,
    author = {Hinkley, David V.},
    title = {Inference about the change-point in a sequence of random variables},
    journal = {Biometrika},
    volume = {57},
    number = {1},
    pages = {1-17},
    year = {1970},
    month = {04},
    doi = {10.1093/biomet/57.1.1},
}

@article{gomes2017adaptive,
  title={Adaptive random forests for evolving data stream classification},
  author={Gomes, Heitor M and Bifet, Albert and Read, Jesse and Barddal, Jean Paul and Enembreck, Fabr{\'\i}cio and Pfharinger, Bernhard and Holmes, Geoff and Abdessalem, Talel},
  journal={Machine Learning},
  volume={106},
  number={9},
  pages={1469--1495},
  year={2017},
  publisher={Springer},
  doi={https://doi.org/10.1007/s10994-017-5642-8}
}

@inbook{kstest,
author = {Berger, Vance W. and Zhou, YanYan},
publisher = {John Wiley \& Sons, Ltd},
isbn = {9781118445112},
title = {Kolmogorov–Smirnov Test: Overview},
booktitle = {Wiley StatsRef: Statistics Reference Online},
chapter = {},
pages = {},
doi = {https://doi.org/10.1002/9781118445112.stat06558},
year = {2014},
}

@article{azzimonti2019reducing,
  title={Reducing probes for quality of transmission estimation in optical networks with active learning},
  author={Azzimonti, Dario and Rottondi, Cristina and Tornatore, Massimo},
  journal={Journal of Optical Communications and Networking},
  volume={12},
  number={1},
  pages={A38--A48},
  year={2019},
  publisher={Optical Society of America},
  doi={https://doi.org/10.1364/JOCN.12.000A38}
}

@article{shayesteh2022automated,
  title={Automated concept drift handling for fault prediction in edge clouds using reinforcement learning},
  author={Shayesteh, Behshid and Fu, Chunyan and Ebrahimzadeh, Amin and Glitho, Roch H},
  journal={IEEE Transactions on Network and Service Management},
  volume={19},
  number={2},
  pages={1321--1335},
  year={2022},
  publisher={IEEE},
  doi={10.1109/TNSM.2022.3153279}
}

@article{moiz2025data,
  title={From data to decision: a multi-stage framework for class imbalance mitigation in optical network failure analysis},
  author={Moiz Ali, Yousuf and Prilepsky, Jaroslaw E and Sambo, Nicola and Pedro, Jo{\~a}o and Hosseini, Mohammad M and Napoli, Antonio and Turitsyn, Sergei K and Freire, Pedro},
  journal={Journal of Optical Communications and Networking},
  volume={18},
  number={1},
  pages={42--58},
  year={2026},
  publisher={Optica Publishing Group},
  doi={https://doi.org/10.1364/JOCN.576774}
}

@article{musumeci2025failure,
  title={Failure management in optical networks with {ML}: a tutorial on applications, challenges, and pitfalls},
  author={Musumeci, Francesco and Tornatore, Massimo},
  journal={Journal of Optical Communications and Networking},
  volume={17},
  number={8},
  pages={C144--C155},
  year={2025},
  publisher={Optica Publishing Group},
  doi={https://doi.org/10.1364/JOCN.551910}
}

@article{bebawi2018comprehensive,
  title={A comprehensive study on EDFA characteristics: temperature impact},
  author={Bebawi, John A and Kandas, Ishac and El-Osairy, Mohamed A and Aly, Moustafa H},
  journal={Applied Sciences},
  volume={8},
  number={9},
  pages={1640},
  year={2018},
  DOI = {10.3390/app8091640},
  publisher={MDPI}
}

@article{andre2005chromatic,
  title={Chromatic dispersion fluctuations in optical fibers due to temperature and its effects in high-speed optical communication systems},
  author={Andr{\'e}, Paulo S and Pinto, Armando N},
  journal={Optics Communications},
  volume={246},
  number={4-6},
  pages={303--311},
  year={2005},
  doi = {https://doi.org/10.1016/j.optcom.2004.11.017},
  publisher={Elsevier}
}
\vspace{-4mm}

\end{document}